\documentclass[conference]{IEEEtran}
\IEEEoverridecommandlockouts
\usepackage{cite}
\usepackage{amsmath,amssymb,amsfonts}
\usepackage{algorithmic}
\usepackage{graphicx}
\usepackage{textcomp}
\usepackage{xcolor}
\usepackage{subcaption}
\def\BibTeX{{\rm B\kern-.05em{\sc i\kern-.025em b}\kern-.08em
    T\kern-.1667em\lower.7ex\hbox{E}\kern-.125emX}}
\begin{document}

\title{3D Point Cloud Feature Explanations Using Gradient-Based Methods\\
\thanks{A. Gupta is funded by the President's Doctoral Scholarship from the University of Manchester and the ACM SIGHPC/Intel Computational and Data Science Fellowship. }
}

\author{\IEEEauthorblockN{Ananya Gupta, Simon Watson, Hujun Yin}
\IEEEauthorblockA{ Department of Electrical and Electronic Engineering \\
The University of Manchester \\
Manchester, UK \\
\{ananya.gupta, simon.watson, hujun.yin\} @manchester.ac.uk
}
}
% \and
% \IEEEauthorblockN{Simon Watson}
% \and
% \IEEEauthorblockN{Hujun Yin}
\maketitle
\thispagestyle{plain}
\pagestyle{plain}
\begin{abstract}
Explainability is an important factor to drive user trust in the use of neural networks for tasks with material impact. However, most of the work done in this area focuses on image analysis and does not take into account 3D data. We extend the saliency methods that have been shown to work on image data to deal with 3D data. We analyse the features in point clouds and voxel spaces and show that edges and corners in 3D data are deemed as important features while planar surfaces are deemed less important. The approach is model-agnostic and can provide useful information about learnt features. Driven by the insight that 3D data is inherently sparse, we visualise the features learnt by a voxel-based classification network and show that these features are also sparse and can be pruned relatively easily, leading to more efficient neural networks. Our results show that the Voxception-ResNet model can be pruned down to 5\% of its parameters with negligible loss in accuracy.
\end{abstract}

\section{Introduction}

Deep neural network (DNN) models are increasingly being used in a number of fields from medical diagnosis~\cite{Esteva2019} to autonomous driving~\cite{Ramos2017} due to their ability to learn meaningful abstractions from data and their successes in many vision tasks. Such models were initially treated as black box operators, but as their popularity has increased, so has the need to make these models interpretable and explainable~\cite{Xiao2018,Lakkaraju2017,Cadamuro2016}.

Explainability is important to gain user trust in areas such as medical diagnosis where machine learning is being used for applications such as cancer prediction~\cite{Xiao2018}. Interpretations are also important for identifying biases in models~\cite{Lakkaraju2017} and can be used for extracting insights and debugging models~\cite{Cadamuro2016}. Driven by these reasons, there has been a lot of work done on the interpretability and explainability of DNNs for image based tasks, and to a lesser extent, language models. We refer readers to \cite{Montavon2018} for a more detailed review on methods for interpretability.

\textit{Interpretability} can be defined as the degree to which a human can understand the cause of a decision. It is the mapping of an abstract concept such as a model's parameters into a domain that can be understood by humans~\cite{Montavon2018}. An example of this would be feature optimisation where given an output neuron, the input image is optimised such that the activation of said neuron would be maximised~\cite{Simonyan2013}. 

\textit{Explainability} is a closely related topic to interpretability. Whereas interpretability focuses on abstract concepts, explainability is the identification of relevant features in the interpretable domain that are useful for attaining a specific decision such as identifying the input pixels that are important for the decision of a classification algorithm. A large number of explainability approaches are gradient-based and produce sensitivity maps or saliency maps~\cite{Simonyan2013,Springenberg2014,Sundararajan2017}. These two terms are used interchangeably in literature, but for the purposes of this work, we will assume the definition given here.

\textit{Saliency maps} in computer vision are used to represent the most noticeable pixels in an image~\cite{Itti1998}. In the context of model explainability, saliency maps denote the pixels that are deemed important for the decision of the model under consideration~\cite{Simonyan2013}. 

% These methods can be broadly divided into feature visualisation and saliency attribution, with some other work being based on dimensionality reduction(t-sne).

Features learnt from 2D data can be visualised and intuited as images~\cite{Zeiler2014}. However, 3D data is not necessarily as intuitively understood. In this work, we explore features learnt by 3D networks as a means of explainability for such networks. More specifically, our contributions are as follows:

\begin{itemize}
\item Methods developed for obtaining saliency maps from image data are extended to deal with 3D point cloud and voxel data.
\item This is the first work that analyses input features that are deemed important to 3D classification networks. 
\item The filters learnt by a 3D voxel-based network are visualised and it is shown that they are inherently sparse and can be pruned efficiently with minimal loss in accuracy, leading to a smaller, more efficient network.   
\end{itemize}

\subsection{Models and Data Types}
3D data can be represented in a number of formats such as point clouds, wireframes, surface models and solids. For the purposes of this study, we limit our focus and experiments to point cloud data\footnote{The kind of data obtained from laser scanners.} and voxel data. 

\begin{figure}
    \centering
    \begin{subfigure}[b]{0.23\textwidth}
    \includegraphics[width = \textwidth]{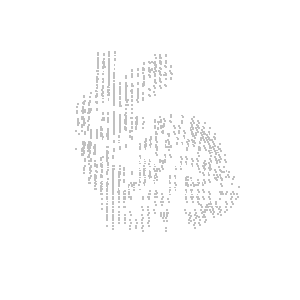} 
    \end{subfigure} 
    \begin{subfigure}[b]{0.23\textwidth}
    \includegraphics[width = \textwidth]{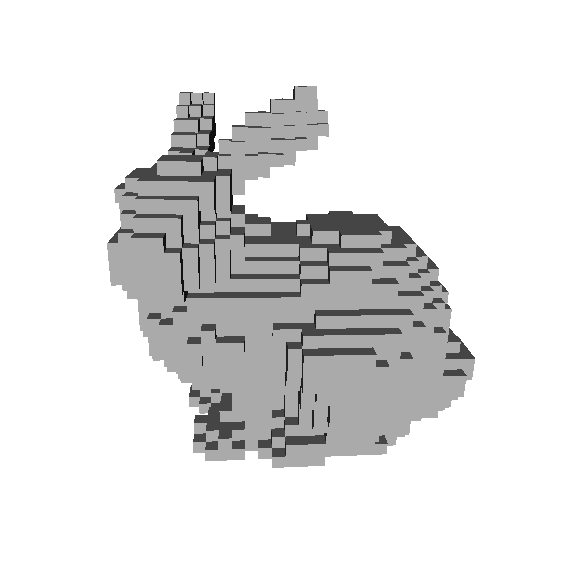} 
    \end{subfigure} 
    \caption{Stanford Bunny~\cite{stanford_bunny}. \textit{Left}: Point cloud representation. \textit{Right}: Voxel representation}
    \label{fig:stanford_bunny}
\end{figure}

Point clouds obtained from LiDAR scanners are unordered point sets with non-uniform density. The point density depends on the sensor scanning pattern and the distance of the surface being scanned from the sensor head. These point clouds can be converted into a uniform voxel format. Voxels are 3D equivalents of pixels, where the space under consideration is divided into a 3D grid and each volumetric element of the grid is known as a voxel. Voxels can be seen as a special case of point clouds with uniform density and quantised dimensions. An example of these two  representations are shown in Fig. \ref{fig:stanford_bunny}.

We choose popular classification models designed for these data types for further investigation: Voxception-ResNet(VRN)~\cite{Brock2016} for voxel data and PointNet++~\cite{Qi2017a} for point cloud data.

\section{Related Work}

\subsection{Explainability Methods}
\label{sec: related_work_explainability}

Explainability is a fast expanding area of research with a number of different sub-areas. Popular approaches to explainability of DNN models include creating a saliency map to identify and highlight the important areas in the input space~\cite{Ancona2017} and creating a proxy model which has similar behaviour to the original model but is easier to explain~\cite{Gilpin2018}.

Perturbation methods such as LIME~\cite{Ribeiro2016}, IME~\cite{Strumbelj2009} and EXPLAIN~\cite{Robnik-Sikonja2008} are often used to create proxy models~\cite{Robnik-Sikonja2018}. These methods are model-agnostic and usually perturb the neighbourhood of an input space to observe the effect of the perturbations on the output. EXPLAIN and IME are based on the premise that ''hiding" some feature or a set of features in the input space can be used to identify the contribution of the aforementioned features to the decision process. EXPLAIN computes the contribution of each feature individually, which has the disadvantage of missing connections between input variables. IME deals with this issue by computing the importance of all subsets of the feature space. However, this leads to the issue of exponential time complexity. 

LIME explains the prediction of a classifier by approximating it with a locally interpretable model around the prediction. It presents the interpretation as an optimisation problem and hence avoids the exponential time complexity issue. An occlusion-based approach was also popularised by Zeiler and Fergus~\cite{Zeiler2014} where  parts of the input were masked and the output decision was computed on a number of such inputs to obtain the importance of a specific input feature. However, similar to the methods described previously, this method was very slow especially as the input space grew large.

Saliency mapping methods are often used for attribution analysis~\cite{Ancona2017}. They are typically gradient-based and are relatively straightforward to compute using backpropagation. They are faster than perturbation-based methods, which typically require a single forward and backward pass through the network. The gradient of the output class score with respect to the input pixels can be visualised as a heatmap where the highest gradient gives the most important pixel since the least change in that pixel would cause the largest change in the output value~\cite{Simonyan2013}.  A number of different techniques such as Guided Backpropagation~\cite{Springenberg2014} and Integrated Gradients~\cite{Sundararajan2017} build on this premise and have some differences in how they propagate gradients which are detailed further in Section \ref{sec:saliency_maps}. These methods have been used for further analysis of neural networks for 3D data since, as pointed out in \cite{Sundararajan2017}, they are immediately applicable to existing models and provide intuitive explanations. 

There are a number of other backpropagation methods. Layerwise Relevance Propagation~\cite{Montavon2019}  was shown to be equivalent within a scaling factor to the element-wise product of the gradient and input~\cite{Shrikumar2017}.  DeepLift~\cite{Shrikumar2017a} assigns an attribution to each input feature based on the relative activation of a reference input. Deep Taylor Decomposition~\cite{Montavon2017} produces sparse explanations but assumes no negative evidence, only showing positive attributions which is not necessarily a valid assumption~\cite{Ancona2017}.

% Decision trees~\cite{Ribeiro}-

% rule extraction

% disentangled representations, features

% salience mapping

\subsection{3D Feature Analysis}
There has been limited related work on analysing 3D features. Some previous work on voxel classification visualised the average surfaces learnt by certain neurons of their model and showed that the initial layers of their model activated mostly on simple surfaces and corners while later layers had high responses for more complex shapes~\cite{Wu2015}. The authors of PointNet++ visualised point cloud patterns learnt by the initial neurons in their network by searching for points in a unit sphere that activated the neurons the most~\cite{Qi2017a}. 

FoldingNet~\cite{Yang2018FoldingNet} was designed as an interpretable model for unsupervised learning where a 2D grid was folded onto a 3D object surface for reconstruction. The authors expressed this as an intrepretable model since the folding could be seen as a granular warping.

% \section{Methods and Related Work}

% The voxel resolution can be limited by the 

% volumetric data can be broadly divided in two formats, voxel-based occupancy grids and point clouds. Both of these formats have their limitations and 3D classification models

\section{Attribution Maps}
\label{sec:saliency_maps}
% Gradient-based methods are commonly used for attribution analysis. They are relatively straightforward to compute using backpropagation. 

% The gradient of the output class score with respect to the input pixels can be visualised as a heatmap where the highest gradient gives the most important pixel since the least change in that pixel would cause the largest change in the output value. 

The formulation for vanilla gradients is given by Equation \ref{eq:saliency_map}. These gradients can be visualised as a heatmap or a saliency map~\cite{Simonyan2013} and are similar to the output from deconvolutional networks~\cite{Zeiler2014}. 

\begin{equation}
Grad_i = \frac{\partial F(x)}{\partial x_i}
\label{eq:saliency_map}    
\end{equation}

The input is given by \(x\) and each element of the input is indexed by subscript \(i\). \(Grad_i\) is the gradient attribution of element \(x_i\) and \(F\) is the function of the neural network.

\textbf{Saliency maps} zero out gradients during the backward pass if the inputs coming into the rectified linear units (ReLU) during the forward pass are negative. On the other hand, \textbf{deconvolutional networks} zero out gradients from the ReLU during the backward pass only if those incoming gradients during the backward are negative.  

\textbf{Guided Backpropagation}~\cite{Springenberg2014} combines the approaches from saliency maps and deconvolutional networks. In this method, the gradient is backpropagated through a ReLU only if the ReLU is switched on (input is non-negative) and the gradient during backward propagation is also non-negative.

Such saliency maps have a lot of noise and a number of methods have been proposed to refine them. A straightforward method to improve the sharpness of the attribution map is to use the element-wise product of the gradient and the input~\cite{Shrikumar2017}.

\textbf{Integrated Gradients}~\cite{Sundararajan2017} computes the average of all the gradients along the straight line path between a baseline, \({x'}\), and the input, \(x\), given by Equation \ref{eq:int_grad}. In the case of an image, the baseline can be a zero image. This method has the desirable property of \textit{completeness}~\cite{Sundararajan2017}, which implies that the attributions add up to the difference between the target and the baseline outputs.

\begin{equation}
IntGrad_i = (x_i - x'_i) \cdot \int_{\alpha=0}^{1}\frac{\partial F(x'+\alpha(x-x'))}{\partial x_i} \partial \alpha
\label{eq:int_grad}    
\end{equation}

% \textbf{Other explainability methods}

% perturbation based methods
% grad cam
% deep lift
% smooth grad

\section{ 3D Features and Network Pruning}

Learnt voxel features can be visualised as 3D filter maps. Since 3D spaces are inherently sparse, we hypothesise that discriminative features for voxel-based networks should also be sparse. However, some 3D CNNs are dense extensions of 2D networks for 3D structures and do not take into account the sparse nature of 3D data. Hence, we took inspiration from pruning methods to test the sparse nature of dense 3D networks.

Pruning methods are broadly divided into fine-grained and coarse-grained pruning~\cite{Cheng2018}. The former is based on pruning individual weights to make the DNNs sparse, while the latter is based on pruning entire kernels or channels. We have extended a popular fine-grained pruning method called Dynamic Network Surgery (DNS)~\cite{Guo2016a} to work with 3D filters to test our hypothesis. The formulation of this pruning method is given below.

%  As part of the results given in Section {FIXME}, we visualise the features learnt by VoxceptionResNet. 

% Driven by the insight that 3D data is inherently sparse and 

The weight tensor representing the weights in layer \(k\) is given by \(W_k\). An additional tensor \(T_k\) is defined which has the same dimensionality as \(W_k\) and is a binary mask matrix to indicate if the corresponding weights in \(W_k\) have been pruned or not. 

The optimization problem is summarised as :
\begin{equation}
\label{eq:prune_opt}
\min_{W_k, \: T_k} \textup{L}(W_k \circ T_k)\; s.t.\: T_k =\: \mathbf{h}_k(W_k),\; 
\end{equation}

where \(\textup{L}\) is the loss function, \( \circ \) represents the Hadamard product. The function \(\textbf{h}_k\) is used to determine the importance of the weights. In our experiments, following the work in \cite{Guo2016a}, \(\textbf{h}_k\) is the absolute value of the weights. Hence, the smaller the absolute value, the less important the weight parameter.

Hence, Equation \ref{eq:prune_opt} looks to minimize the loss by optimising the values of \(W_k\) and \(T_k\) and  is an N.P. hard problem. In this case, these values are optimised iteratively where the weight updates are given by a slight modification of the standard gradient descent algorithm during backpropagation in order to incorporate the weight mask as follows:

\begin{equation}
W_k \leftarrow W_k - \; \beta \frac{\partial}{\partial(W_kT_k) } \textup{L}(W_k \circ T_k),\; 
\end{equation}
where \(\beta\) represents the learning rate. 

This update carries through for all weights, including the ones where the corresponding value in the weight mask is zero, allowing the weight mask to be updated by removing certain values and restoring others during the next forward pass operation as follows:

\begin{equation}
\mathbf{h}_k(W_k) = \left\{\begin{matrix}
0 & if \; t_k > \lvert W_k\rvert \\ 
1 & if \; t_k < \lvert W_k\rvert
\end{matrix}\right.
\end{equation}
 where the threshold \(t_k\) is defined using the mean and variance of the absolute values of the weights in layer \(k\).
% 3D CNNs are dense expansions of 2D CNNs

% Different methods have been proposed in order to increase neural network efficiency. These include student-teacher network distillation, low-rank matrix factorisation and pruning.

\section{Experimental Details}
 
The Pointnet++ and VRN models were trained according to the details given by the original authors of the respective papers. The VRN model was reimplemented in Pytorch where the original implementation of Pointnet++ was used for all experiments.

Following the implementation in the original papers, the Modelnet40 models were voxelised to a resolution of 32x32x32 for VRN and 1024 points were sampled on the surface of each model for Pointnet++.

The baseline was assumed to be an empty voxel space for integrated gradients, with 50 steps between the baseline and the input.

\section{Results}

\subsection{Attribution Maps}

\begin{figure*}[h!]
\begin{tabular}{|c|c|c|c|}
\hline 
Point Cloud & Vanilla Grad & Guided Backprop & Integrated Grad \\ \hline
\begin{subfigure}[b]{0.23\textwidth}
\includegraphics[width = \textwidth]{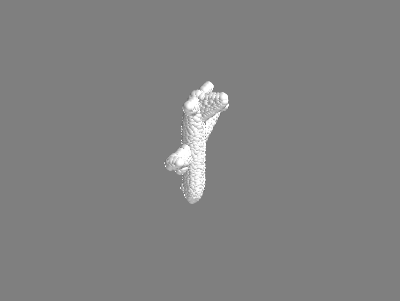} 
\end{subfigure} &
\begin{subfigure}[b]{0.23\textwidth}
\includegraphics[width = \textwidth]{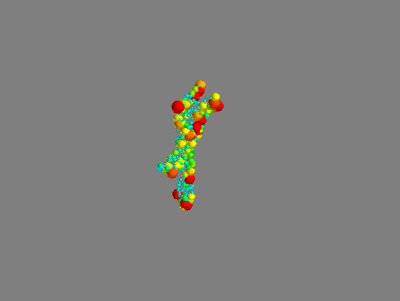} 
\end{subfigure} &
\begin{subfigure}[b]{0.23\textwidth}
\includegraphics[width = \textwidth]{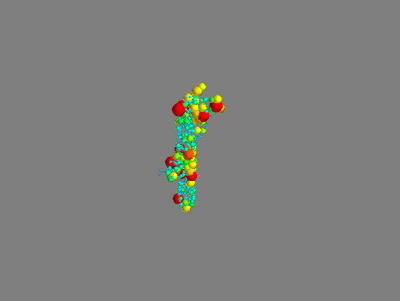} 
\end{subfigure} &
\begin{subfigure}[b]{0.23\textwidth}
\includegraphics[width = \textwidth]{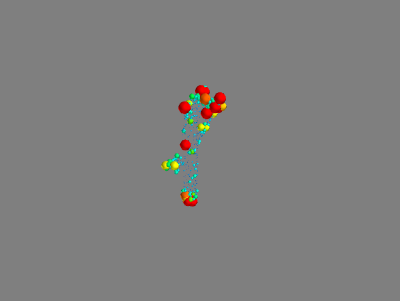} 
\end{subfigure} \\
\multicolumn{4}{|c|}{Airplane} \\ \hline
\begin{subfigure}[b]{0.23\textwidth}
\includegraphics[width = \textwidth]{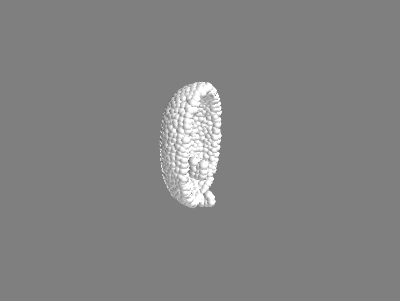} 
\end{subfigure} &
\begin{subfigure}[b]{0.23\textwidth}
\includegraphics[width = \textwidth]{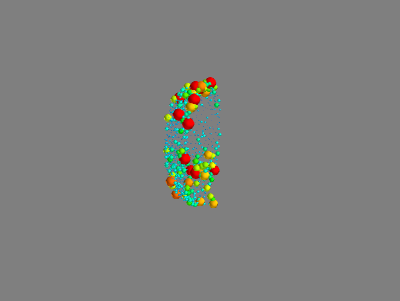} 
\end{subfigure} &
\begin{subfigure}[b]{0.23\textwidth}
\includegraphics[width = \textwidth]{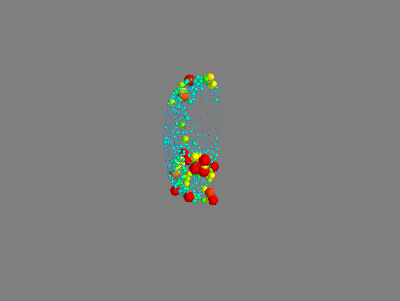} 
\end{subfigure} &
\begin{subfigure}[b]{0.23\textwidth}
\includegraphics[width = \textwidth]{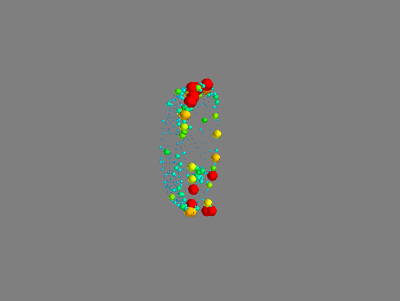} 
\end{subfigure} \\
\multicolumn{4}{|c|}{Bathtub}\\ \hline
\begin{subfigure}[b]{0.23\textwidth}
\includegraphics[width = \textwidth]{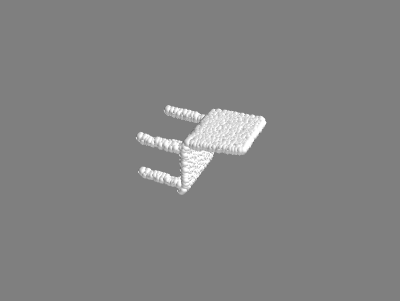} 
\end{subfigure} &
\begin{subfigure}[b]{0.23\textwidth}
\includegraphics[width = \textwidth]{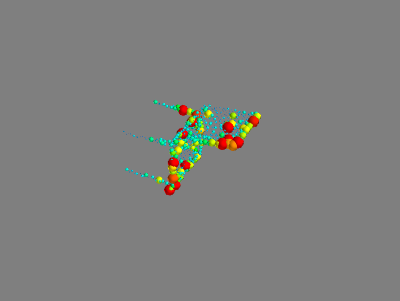} 
\end{subfigure} &
\begin{subfigure}[b]{0.23\textwidth}
\includegraphics[width = \textwidth]{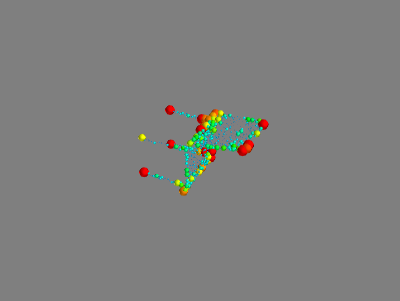} 
\end{subfigure} &
\begin{subfigure}[b]{0.23\textwidth}
\includegraphics[width = \textwidth]{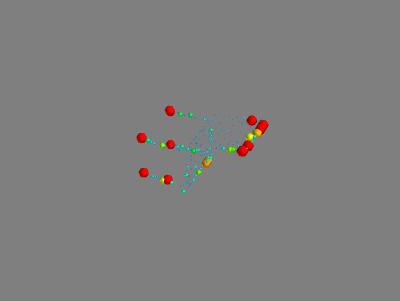} 
\end{subfigure} \\
\multicolumn{4}{|c|}{Chair} \\ \hline
\begin{subfigure}[b]{0.23\textwidth}
\includegraphics[width = \textwidth]{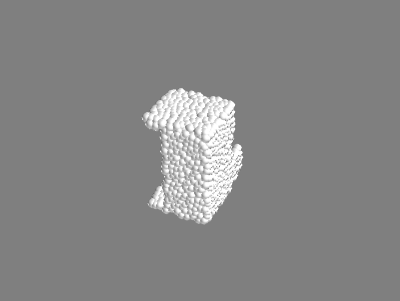} 
\end{subfigure} &
\begin{subfigure}[b]{0.23\textwidth}
\includegraphics[width = \textwidth]{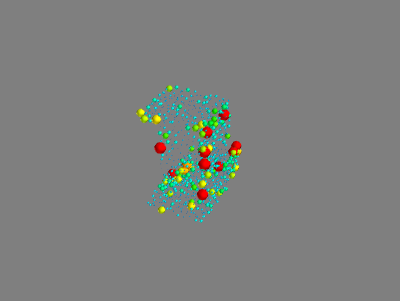} 
\end{subfigure} &
\begin{subfigure}[b]{0.23\textwidth}
\includegraphics[width = \textwidth]{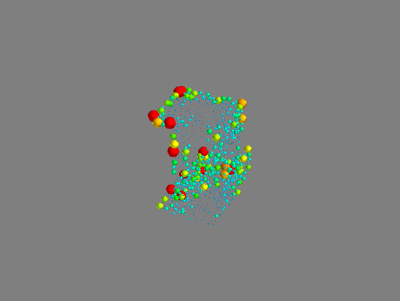} 
\end{subfigure} &
\begin{subfigure}[b]{0.23\textwidth}
\includegraphics[width = \textwidth]{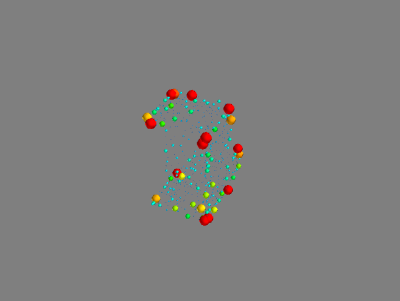} 
\end{subfigure} \\
\multicolumn{4}{|c|}{Desk}\\ \hline
\begin{subfigure}[b]{0.23\textwidth}
\includegraphics[width = \textwidth]{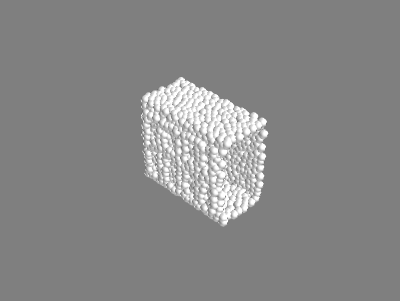} 
\end{subfigure} &
\begin{subfigure}[b]{0.23\textwidth}
\includegraphics[width = \textwidth]{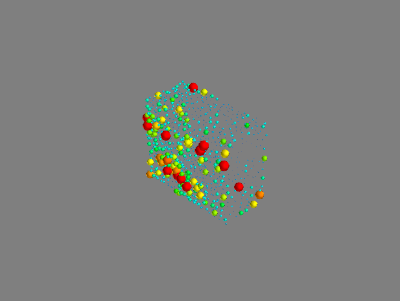} 
\end{subfigure} &
\begin{subfigure}[b]{0.23\textwidth}
\includegraphics[width = \textwidth]{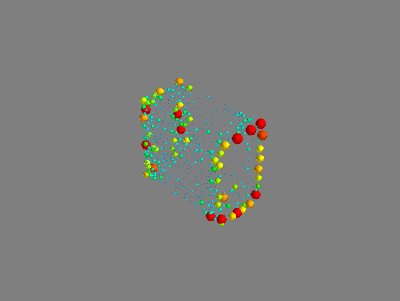} 
\end{subfigure} &
\begin{subfigure}[b]{0.23\textwidth}
\includegraphics[width = \textwidth]{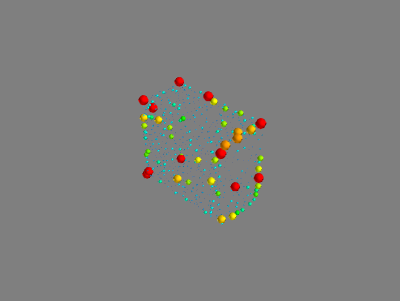} 
\end{subfigure} \\
\multicolumn{4}{|c|}{Dresser}
% \\ \hline
% \begin{subfigure}[b]{0.23\textwidth}
% \includegraphics[width = \textwidth]{images/pointnet/p_monitor_orig.png} 
% \end{subfigure} &
% \begin{subfigure}[b]{0.23\textwidth}
% \includegraphics[width = \textwidth]{images/pointnet/p_monitor_vanilla.png} 
% \end{subfigure} &
% \begin{subfigure}[b]{0.23\textwidth}
% \includegraphics[width = \textwidth]{images/pointnet/p_monitor_guided.png} 
% \end{subfigure} &
% \begin{subfigure}[b]{0.23\textwidth}
% \includegraphics[width = \textwidth]{images/pointnet/p_monitor_int.png} 
% \end{subfigure} \\
% \multicolumn{4}{|c|}{Monitor}
\\ \hline
\begin{subfigure}[b]{0.23\textwidth}
\includegraphics[width = \textwidth]{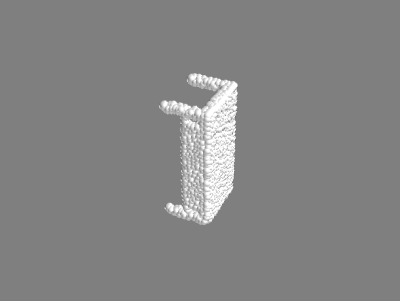} 
\end{subfigure} &
\begin{subfigure}[b]{0.23\textwidth}
\includegraphics[width = \textwidth]{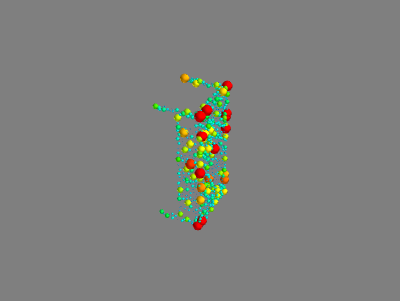} 
\end{subfigure} &
\begin{subfigure}[b]{0.23\textwidth}
\includegraphics[width = \textwidth]{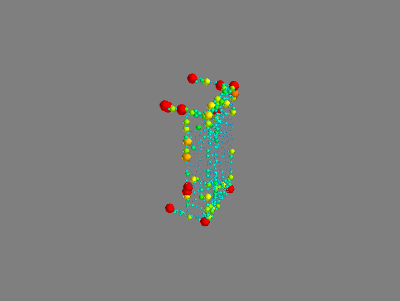} 
\end{subfigure} &
\begin{subfigure}[b]{0.23\textwidth}
\includegraphics[width = \textwidth]{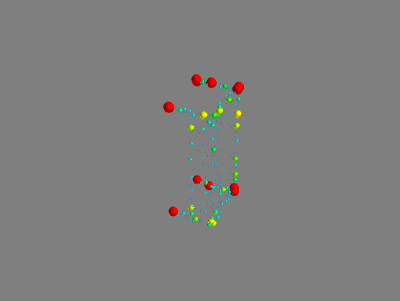} 
\end{subfigure}\\
\multicolumn{4}{|c|}{Table}
\\ \hline
\end{tabular}
\caption{Visualisation of attribution maps for Pointnet++. The attributions are given as a heatmap by Red (large) to Blue (small). }
\label{fig:pointnet_saliency_maps}
\end{figure*}

Examples of attribution maps for Pointnet++ obtained using vanilla gradients, guided backpropagation and integrated gradients as outlined in Section \ref{sec:saliency_maps} are shown in Figure \ref{fig:pointnet_saliency_maps}. As can be seen from the figure, vanilla gradients attribute more importance to edges and corners than they do to flat surfaces, though the relevance of points along surfaces is not uniform, leading to the assumption that these attribution maps are fairly noisy. 

The maps obtained using guided backpropagation are somewhat more uniform, with higher saliency attributions given to highly discriminative features, such as the stand in the case of a television and the tap in the case of a bathtub. The clearest results are achieved with the integrated gradients, which identify corners and edges and do not give much importance to flat surfaces.

\begin{figure*}[h!]
\begin{tabular}{|c|c|c|c|}
\hline 
Point Cloud & Vanilla Grad & Masked Vanilla & Integrated Grad \\ \hline
& & & \\
\begin{subfigure}[b]{0.23\textwidth}
\includegraphics[width = \textwidth]{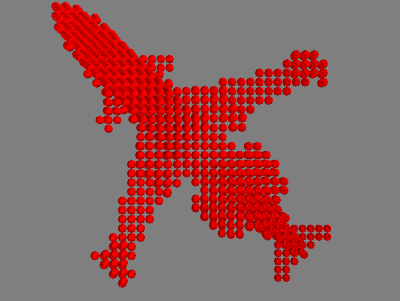} 
\end{subfigure} &
\begin{subfigure}[b]{0.23\textwidth}
\includegraphics[width = \textwidth]{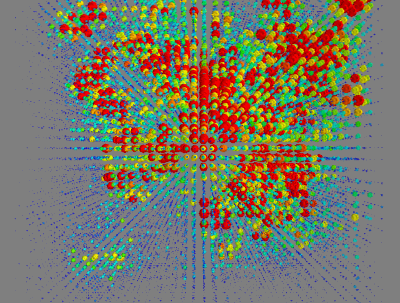} 
\end{subfigure} &
\begin{subfigure}[b]{0.23\textwidth}
\includegraphics[width = \textwidth]{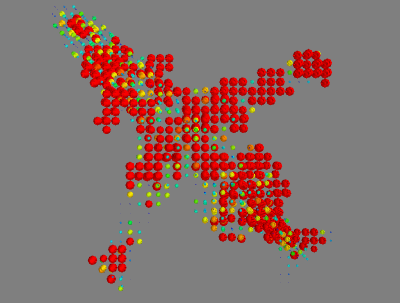} 
\end{subfigure} &
\begin{subfigure}[b]{0.23\textwidth}
\includegraphics[width = \textwidth]{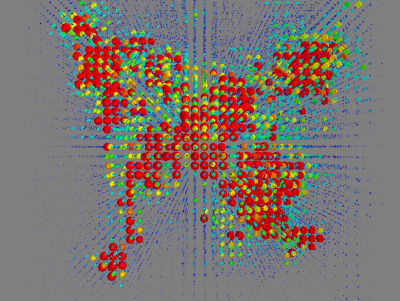} 
\end{subfigure}\\
\multicolumn{4}{|c|}{Airplane}
\\ \hline
\begin{subfigure}[b]{0.23\textwidth}
\includegraphics[width = \textwidth]{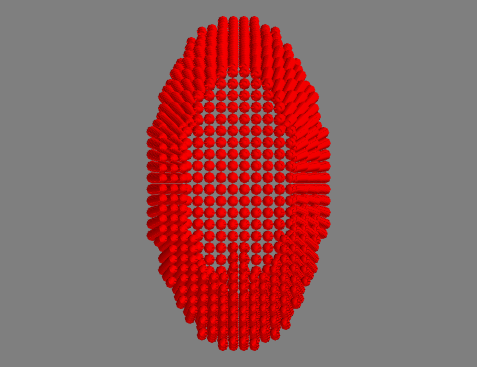} 
\end{subfigure} &
\begin{subfigure}[b]{0.23\textwidth}
\includegraphics[width = \textwidth]{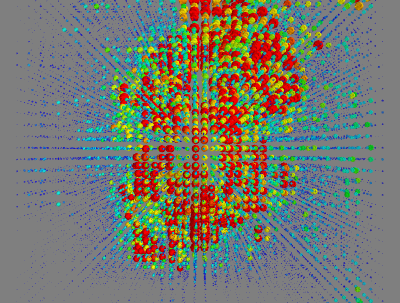} 
\end{subfigure} &
\begin{subfigure}[b]{0.23\textwidth}
\includegraphics[width = \textwidth]{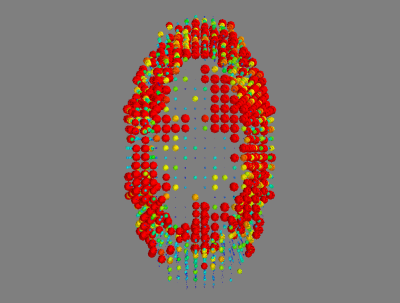} 
\end{subfigure} &
\begin{subfigure}[b]{0.23\textwidth}
\includegraphics[width = \textwidth]{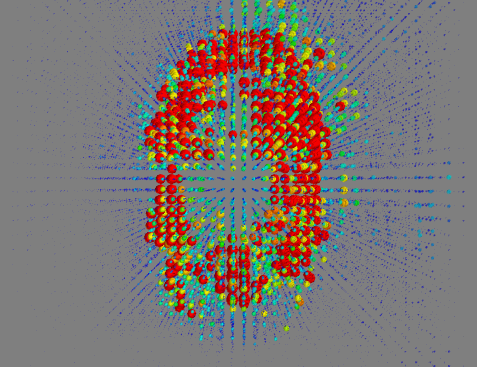} 
\end{subfigure} \\
\multicolumn{4}{|c|}{Bathtub}
\\ \hline
\begin{subfigure}[b]{0.23\textwidth}
\includegraphics[width = \textwidth]{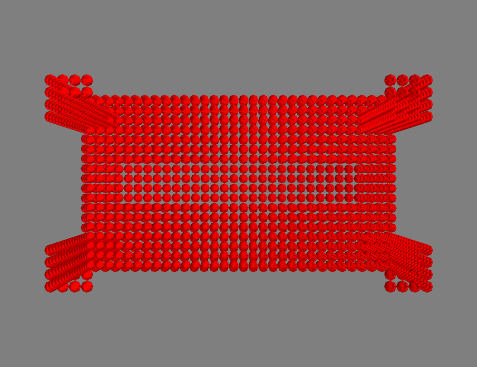} 
\end{subfigure} &
\begin{subfigure}[b]{0.23\textwidth}
\includegraphics[width = \textwidth]{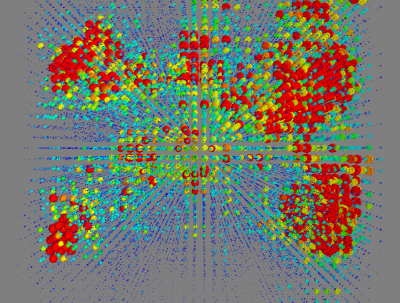} 
\end{subfigure} &
\begin{subfigure}[b]{0.23\textwidth}
\includegraphics[width = \textwidth]{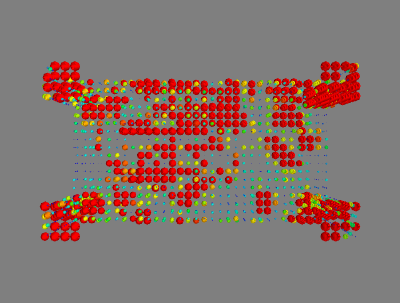} 
\end{subfigure} &
\begin{subfigure}[b]{0.23\textwidth}
\includegraphics[width = \textwidth]{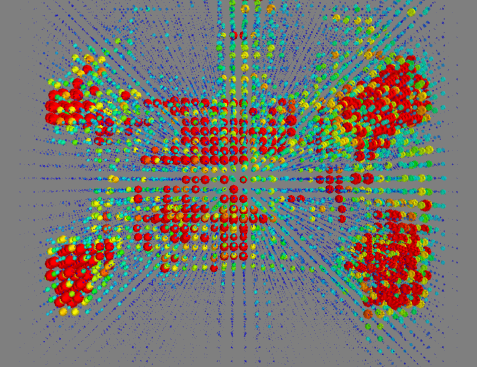} 
\end{subfigure} \\
\multicolumn{4}{|c|}{Bench}
\\ \hline
\begin{subfigure}[b]{0.23\textwidth}
\includegraphics[width = \textwidth]{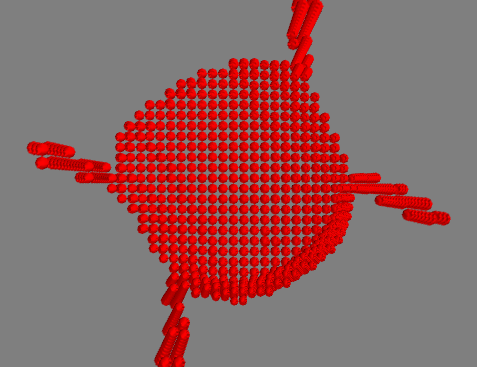} 
\end{subfigure} &
\begin{subfigure}[b]{0.23\textwidth}
\includegraphics[width = \textwidth]{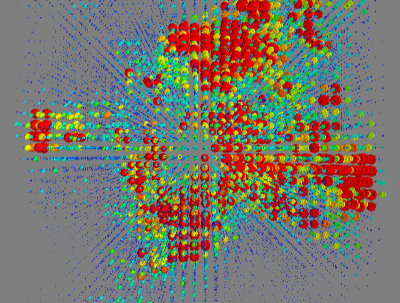} 
\end{subfigure} &
\begin{subfigure}[b]{0.23\textwidth}
\includegraphics[width = \textwidth]{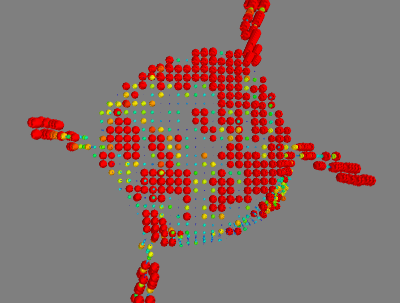} 
\end{subfigure} &
\begin{subfigure}[b]{0.23\textwidth}
\includegraphics[width = \textwidth]{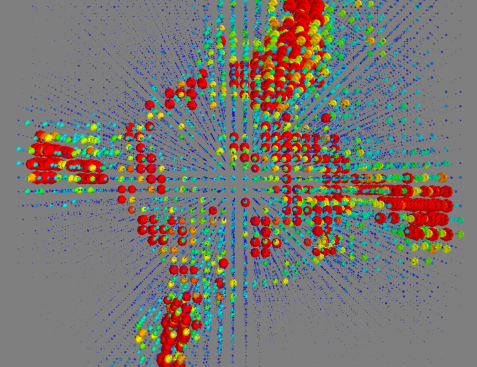} 
\end{subfigure} \\
\multicolumn{4}{|c|}{Chair}
\\ \hline
\begin{subfigure}[b]{0.23\textwidth}
\includegraphics[width = \textwidth]{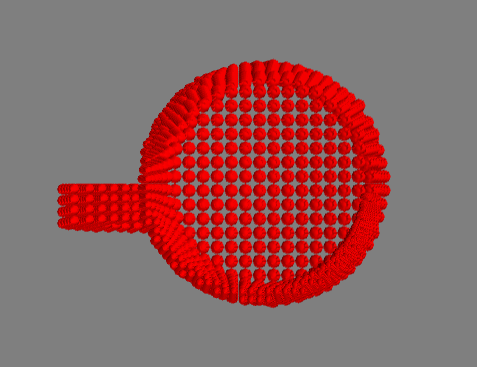} 
\end{subfigure} &
\begin{subfigure}[b]{0.23\textwidth}
\includegraphics[width = \textwidth]{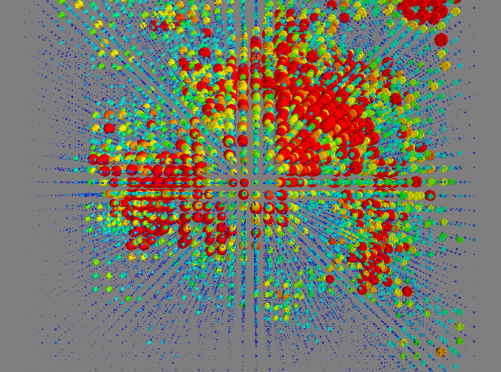} 
\end{subfigure} &
\begin{subfigure}[b]{0.23\textwidth}
\includegraphics[width = \textwidth]{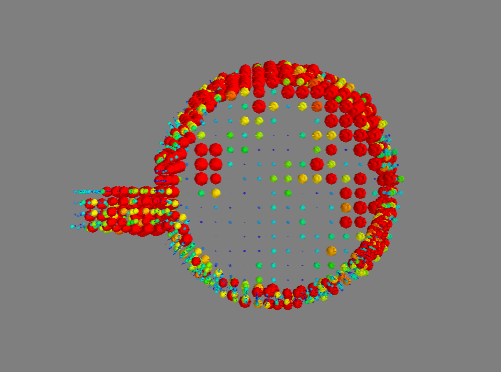} 
\end{subfigure} &
\begin{subfigure}[b]{0.23\textwidth}
\includegraphics[width = \textwidth]{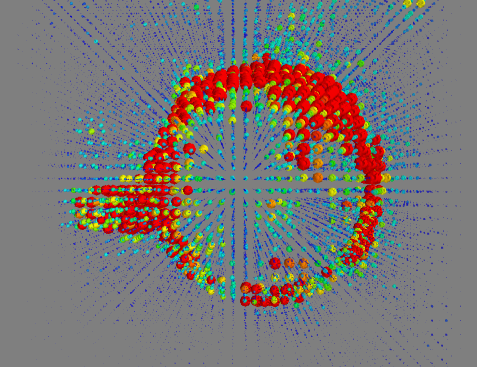} 
\end{subfigure} \\
\multicolumn{4}{|c|}{Cup}
\\ \hline
\end{tabular}
\caption{Visualisation of the attribution maps for VRN.}
\label{fig:vrn_maps}
\end{figure*}

The attribution maps for VRN are shown in Figure \ref{fig:vrn_maps}. As can be seen, the vanilla gradient maps are a lot noisier in this case as compared to those for Pointnet++. This is due to the fact that the voxel inputs encode free space along with occupied space while the point clouds only encode occupied space. Hence, the vanilla gradients in the voxel space are also, in a way, affected by the empty voxels. In order to make these maps less noisy, we show the element-wise product of the gradients and input as the `Masked Vanilla' output in Figure \ref{fig:vrn_maps}.

These masked maps can show the salient features in the input space more clearly. For example, in the case of the cup, the handle and the shape of the cup are important for the classification. It is interesting to compare the masked gradients with the results of the integrated gradients, where the most important voxels seem to overlap. The latter does deem some voxels in the unoccupied space as being important. However, in contrast to vanilla gradients, these unoccupied voxels are given almost negligible importance.

\subsection{Pointnet++ Error Analysis}

\begin{figure*}[h!]
{\includegraphics[width=0.95\textwidth]{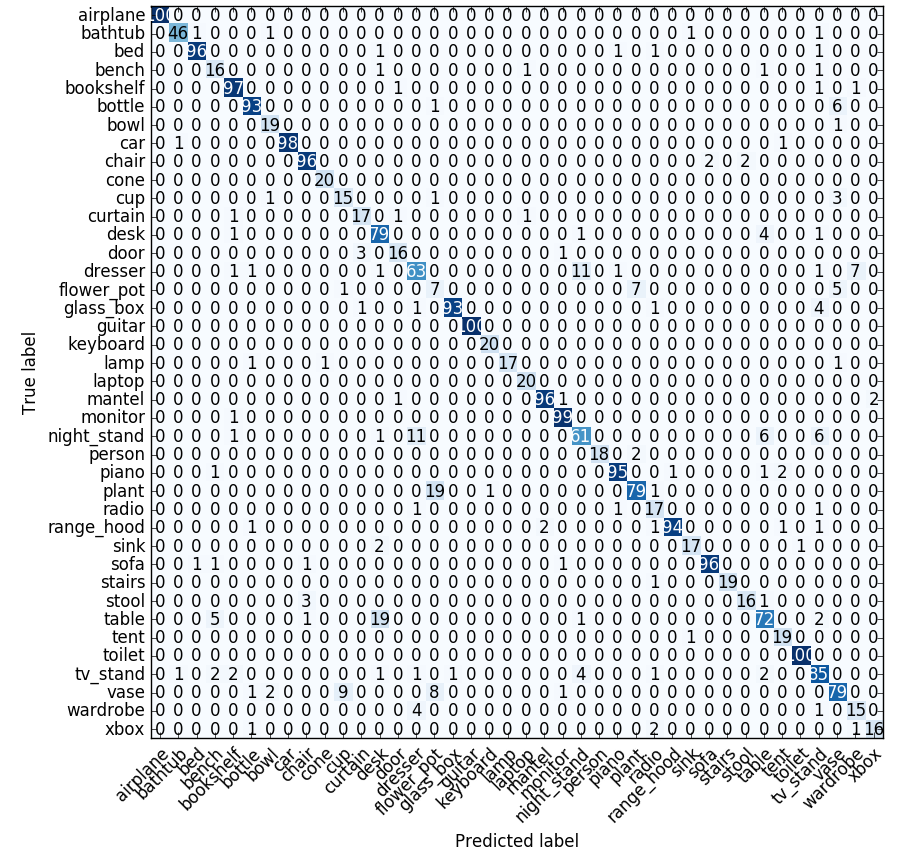}}
\caption[Pointnet++ results]{Confusion matrix of Pointnet++ results on ModelNet40.}
\label{fig:pointnet_conf_mat}
\end{figure*}

The PointNet++ model achieved 90.2\% accuracy on the ModelNet40 test dataset when trained according to the parameters given by the original authors. The confusion matrix for the test set is shown in Figure \ref{fig:pointnet_conf_mat}.

\begin{figure*}[h!]
\begin{tabular}{|c|c|c|c|c|}
\hline 
%  & & & &\\
% Point Cloud & Integrated Grad \\ \hline
Point Cloud &
\begin{subfigure}[b]{0.2\textwidth}
\includegraphics[width = \textwidth]{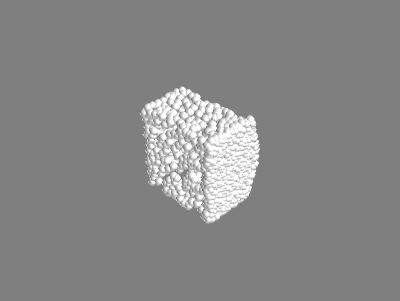} 
\end{subfigure} &
\begin{subfigure}[b]{0.2\textwidth}
\includegraphics[width = \textwidth]{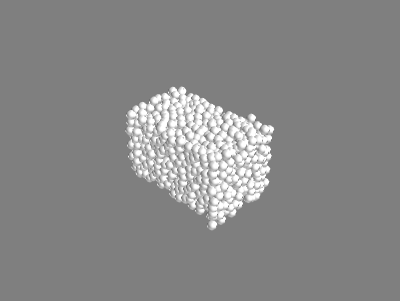} 
\end{subfigure} &
\begin{subfigure}[b]{0.2\textwidth}
\includegraphics[width = \textwidth]{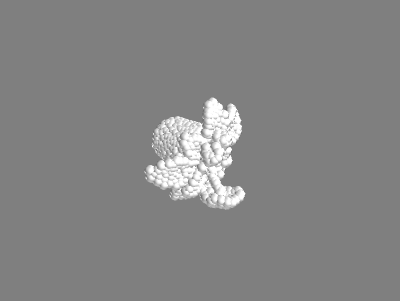} 
\end{subfigure} &
\begin{subfigure}[b]{0.2\textwidth}
\includegraphics[width = \textwidth]{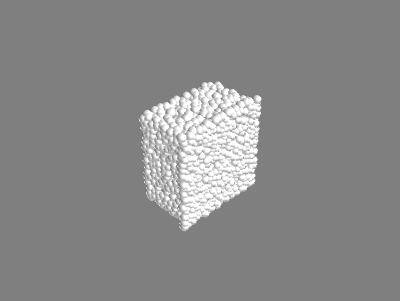} 
\end{subfigure} \\ \hline

% & & & & \\
Integrated Grad & 
\begin{subfigure}[b]{0.2\textwidth}
\includegraphics[width = \textwidth]{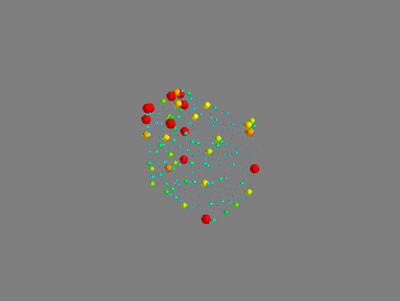} 
\end{subfigure} &
% \multicolumn{2}{|c|}{Dresser identified as Night stand}
% \\ \hline

\begin{subfigure}[b]{0.2\textwidth}
\includegraphics[width = \textwidth]{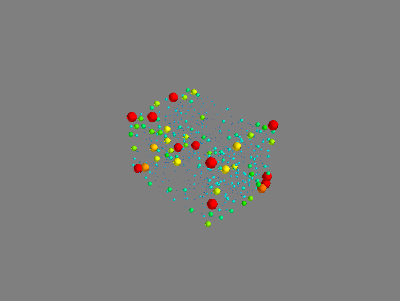} 
\end{subfigure} &
% \multicolumn{2}{|c|}{Night stand identified as Dresser}
% \\ \hline

\begin{subfigure}[b]{0.2\textwidth}
\includegraphics[width = \textwidth]{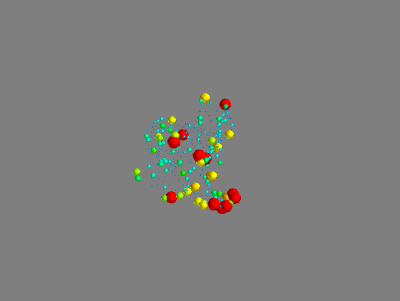}
\end{subfigure} &
% \multicolumn{2}{|c|}{Plant identified as Flowerpot}
% \\ \hline

\begin{subfigure}[b]{0.2\textwidth}
\includegraphics[width = \textwidth]{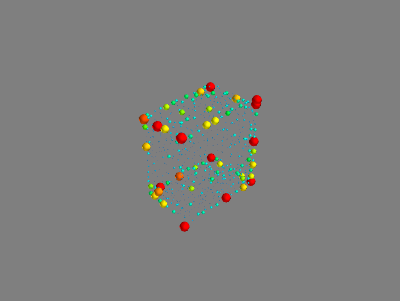} 
\end{subfigure}
% \multicolumn{2}{|c|}{Table identified as Desk}
\\ \hline

Original Class & Dresser & Night Stand & Plant & Table \\ \hline
Classification & Night Stand & Dresser & Flowerpot & Desk \\ \hline
\end{tabular}
\caption{Visualisation of the incorrectly classified point clouds.}
\label{fig:incorrect_pc}
\end{figure*}

The confusion matrix shows that the major errors are between classes that have a fair amount of semantic overlap, such as plants being recognised as flower pots and tables being labelled as desks. Some of these misidentified objects are shown in Figure \ref{fig:incorrect_pc} along with their saliency maps based on the Integrated Gradients. From these images, it can be seen that the mistakes made by the model could have been also made by humans since these classes are fairly similar.

\subsection{Features Learnt by Voxel Networks}

\begin{figure*}[h]
\centering
{\includegraphics[width=0.9\textwidth]{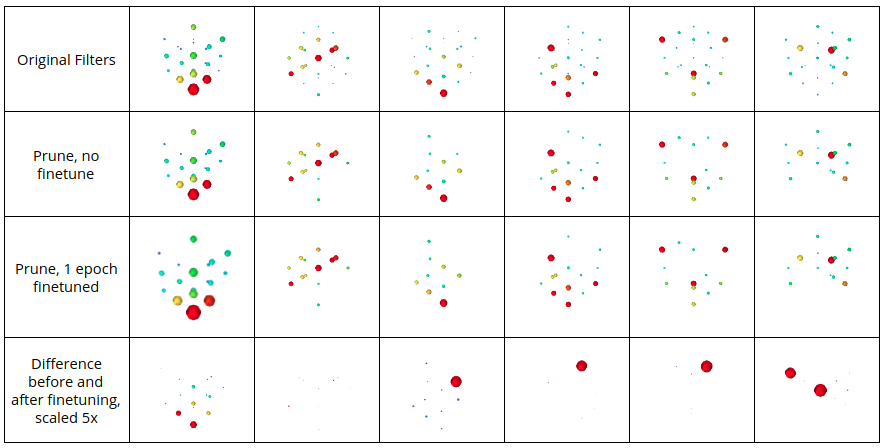}}
\caption[Voxel features]{VRN features where each column denotes one feature from the first layer of the network. From top to bottom, the features are as follows: \textit{top}: original features, \textit{second row}: pruned, \textit{third row}: finetuned, \textit{bottom}: 5x scaled difference between the pruned and finetuned version. It can be seen that there is very little difference between the weights of the pruned network and original network because the learnt features are inherently sparse.}
\label{fig:3d_prune}
\end{figure*}

Some of the features learnt by VRN have been visualised in Figure \ref{fig:3d_prune} where the size of each element denotes the relative absolute value of the weight. The figure also shows the same features after pruning and finetuning. The difference between the pruned features with and without finetuning is minimal and has also been shown.

\begin{table}[]
\centering
\caption{3D Weight Pruning Results for VRN}
\label{tab:3d_pruning}
\begin{tabular}{|l|c|c|c|}
\hline
& \textbf{\# Parameters  }          &  \textbf{Params Left} &   \textbf{ Accuracy}     \\ 
&                                   & (\%)                  &               (\%)    \\ \hline
Original Model        & 13,829,792         & 100         & 87.77   \\ \hline
Prune, no finetune    & 728,092           & 5.26 & 62.39   \\ \hline
Prune, 1 epoch tuning & 700,720           & 5 & 87.18   \\ \hline
\end{tabular}
\end{table}

From the results in Table \ref{tab:3d_pruning}, it can be seen that pruning the network down to almost 5\% of its parameters decreased the accuracy by 25\% but finetuning for only 1 epoch brings the accuracy back up to the original results even with the pruned model. This is contrary to the process with image based models which require finetuning in the order of over 10k iterations~\cite{Guo2016a}. This seems to support the hypothesis that the 3D features learnt are fairly sparse and removal of small weights does not overly affect the performance. The visualisations in Figure \ref{fig:3d_prune} also verify this as it can be seen that the difference between the original model and the pruned and finetuned model is minimal.

\section{Conclusions}

This work is an initial study on explainability of neural network models for 3D data. To this end, popular attribution methods currently used with image data have been extended to deal with point cloud and voxel data. It has also been shown that the features learnt by voxel based networks are sparse and can be pruned easily with little finetuning required.
% In this work, we have analysed the features in 3D data that are deemed important for its classification. This was done by extending

Our results show that edges and corners are considered as important features by gradient-based methods, while planar surfaces do not contribute as much to the classification decision. Vanilla gradients are fairly noisy but the use of integrated gradients makes the attribution maps more uniform. In the case of voxel-based inputs, vanilla gradients attribute a lot of importance to empty space. These attributions become a lot more sensible when masked gradients are used, or with the use of integrated gradients.

We have visualised the learnt features of the voxel classification network and showed the sparsity of these learnt filters. The network can be pruned down to 5\% of its original number of parameters with minimal loss in accuracy and only one epoch of finetuning; as compared to image based networks which require over 10k iterations of iterative pruning and finetuning. We believe this is due to the fact that 3D data is inherently sparse and hence the features learnt for this kind of data are also sparse.

This work can be extended in a number of directions. A natural extension of this work would be to use the insights gained from the gradient-based models to prune DNNs during training time rather than as a post-processing step. Some other relatively straightforward extensions include testing 3D models using some perturbation-based methods such as the ones described in Section \ref{sec: related_work_explainability}. Another important area of research is the systematic quantification of the extracted explanations. We refer readers to \cite{Ancona2017} for ideas on the same. 

\bibliographystyle{IEEEtran}
\bibliography{main}
\end{document}